\documentclass{article}

\usepackage[final,nonatbib]{nips_2017}

\usepackage[utf8]{inputenc} 
\usepackage[T1]{fontenc}    
\usepackage{hyperref}       
\usepackage{url}            
\usepackage{booktabs}       
\usepackage{amsfonts}       
\usepackage{nicefrac}       
\usepackage{microtype}      
\usepackage{amsmath,amsfonts,amssymb,amsthm}
\usepackage{bm}
\usepackage{graphicx}
\usepackage{subfigure}
\usepackage{floatrow}
\usepackage{capt-of}
\usepackage{multirow,array}
\usepackage{color}

\newcommand{\bth}{\bm{\theta}}

\newcommand{\R}{\mathbb{R}}
\newcommand{\bE}{\mathbb{E}}

\newcommand{\RR}{\mathcal{R}}
\newcommand{\CS}{\mathcal{S}}
\newcommand{\NN}{\mathcal{N}}

\newcommand{\BB}{\mathcal{B}}
\newcommand{\CA}{\mathcal{A}}
\newcommand{\HH}{\mathcal{H}}

\newcommand{\bx}{\bm{x}}
\newcommand{\by}{\bm{y}}
\newcommand{\va}{\bm{a}}
\newcommand{\vb}{\bm{b}}
\newcommand{\vc}{\bm{c}}

\newcommand{\bv}{\bm{v}}

\newcommand{\good}{\text{good}}
\newcommand{\bad}{\text{bad}}

\newcommand{\argmin}{\text{argmin}}
\newcommand{\var}{\text{\bf Var}}
\newcommand{\train}{\text{\bf train}}

\newcommand{\attack}{\text{\bf attack}}

\newcommand{\vol}{\text{\bf Vol}}

\newcommand{\half}{\frac{1}{2}}

\newtheorem{theorem}{Theorem}
\newtheorem{corollary}{Corollary}

\title{Towards Understanding Generalization of Deep Learning: Perspective of Loss Landscapes }
\author{
	Lei Wu \\
	School of Mathematics, Peking University\\
	Beijing, China\\
	\texttt{leiwu@pku.edu.cn} \\
	\And
	Zhanxing Zhu \\
	Beijing Institute of Big Data Research (BIBDR)\\
	Center for Data Science, Peking University\\
	Beijing, China\\
	\texttt{zhanxing.zhu@pku.edu.cn}
	\And
	Weinan E \\
	Beijing Institute of Big Data Research (BIBDR) \\
	Center for Data Science and BICMR, Peking University, Beijing, China\\
	Department of Mathematics and PACM, Princeton University, Princeton, NJ, USA\\
	\texttt{weinan@math.princeton.edu}
}

\begin{document}

\maketitle

\begin{abstract}
  It is widely observed that deep learning models with learned parameters generalize well, even with much more model parameters than the number of training samples. We systematically investigate the underlying reasons why deep neural networks often generalize well, and reveal the difference between the  minima (with the same training error) that generalize well and those they don't. We show that it  is the characteristics the landscape of the loss function that explains the good generalization capability. For the landscape of loss function for deep networks,  the volume of basin of attraction of good minima dominates over that of poor minima, which guarantees optimization methods with random initialization to converge to good minima.  We theoretically justify our findings through analyzing 2-layer neural networks; and show that the low-complexity solutions have a small norm of Hessian matrix with respect to model parameters. For deeper networks, extensive  numerical evidence helps to support our arguments.
\end{abstract}

\section{Introduction}
\label{sec:intro}


Recently, deep learning~\cite{lecun2015deep}  has achieved remarkable success in various application areas. In spite  of its powerful modeling capability, we know little about why deep learning works so well from a theoretical perspective. This is widely known as the ``black-box'' nature of deep learning.

One key observation is that,  \emph{most of deep neural networks with learned parameters often generalize very well empirically, even equipped with much more effective parameters than the number of training samples, i.e. high-capacity}.
 According to conventional statistical learning theory (including VC dimension~\cite{vapnik1998statistical} and Rademacher complexity measure~\cite{bartlett2002rademacher}), in such over-parameterized and non-convex models,  the system is easy to get stuck into local
minima that generalize badly. Some regularizations are required to control the generalization error.
 However, as shown in \cite{zhang2016understanding}, high-capacity  neural networks without any regularization can still obtain low complexity solutions and  generalize well; and suitable regularization only helps to improve the test error to a small margin. Thus, statistical learning theory cannot explain the generalization ability of deep learning models.

It is worthy of noting that we call the solutions (or minima) with the same and small training error  ``good '' or ``bad'' if they have \emph{significant difference} of generalization performance, i.e. test accuracy. Take the task of MNIST digit classification as  an example, with the same $100\%$ training accuracy,  we are curious about the  striking difference between the minima achieving above $95\%$ test accuracy and those bad ones like a random guess, rather than the small difference between the solutions above  $95\%$ test accuracy. For those bad solutions that are rarely observed in normal training procedures, we find them by  intentionally adding attacking data to the original training set. To the best of our knowledge, this is the first time that the bad solutions (with the same and small training error as good ones) are accessible, see Section~\ref{sect: contrsuct-minima}. This directly provides the possiblity of considering the difference between the good and bad solutions.


 In this work, we aim to answer two crucial questions towards understanding the generalization of deep learning:
 \begin{itemize}
 \item \textbf{Question 1}: \textit{ What is the property that distinguishes the good solutions (obtained from optimizers) from those that generalize poorly? }
 \item \textbf{Question 2}: \textit{ Although there exist many solutions with bad generalization performance,
 why do the optimizers with random initialization almost surely converge to the minima generalizing well? }
\end{itemize}
We provide reasonable explanation to both of the two questions.

For the first one, we find that local minima with large volume of attractors often lead good generalization performance, as theoretically studied in Section~\ref{sec:lands_and_init} and~\ref{sec:2layer}. A Hessian-based analysis is proposed for quantifying the volume of attractor, based on which extensive numerical evidence (see Section~\ref{sec:exp}) reveals the important relationship between the generalization performance and the volume of basin of attractor.

For the second question, several factors are investigated systematically that might endow neural networks with powerful generalization capability, including initialization, optimization and the characteristics of the landscapes of objective functions. Through extensive theoretical and empirical justifications, we exclude the effects of optimization, as illustrated in Section~\ref{sec:sgd}. And we conjecture that it is the characteristics of the landscapes of loss functions that automatically guarantees the optimization methods with random initialization to converge to good  minima almost surely, see Section~\ref{sec:lands_and_init}. This property is irrelevant with the types of optimization methods adopted during the training.


Our findings dramatically differ from traditional understanding on generalization of neural network, which attribute it to some particular optimizers (e.g. stochastic gradient descent), or certain regularizing techniques (e.g. weight decay or Dropout). These factors can only explain the small difference between the good solutions, rather than the significant difference between good and bad minima. We conjecture that the ``mystery'' of small generalization error is due to the special structure of neural networks.
This is justified through theoretically analyzing the landscape of 2-layer neural network, as shown in Section~\ref{sec:2layer}.

\subsection*{Related work}
Different approaches have been employed to discuss the generalization of neural networks. Some works~\cite{hardt2015train,zhang2016understanding} explored the implicit regularization property of SGD. Another perspective relies on the geometry of loss function around a global minimum. It argues that the solutions that
generalize well lie in flat valleys, while bad ones always are located in the sharp ravine. This observation dates back to the early
works~\cite{hochreiter1995simplifying,hochreiter1997flat}. Recently~\cite{keskar2016large} adopts this perspective to explain why small-batch SGD often converges to
the solution generalizing better than large-batch SGD; and the authors of~\cite{chaudhari2016entropy,ye2017langevin} proposed to use controllable noise to bias the SGD to find flat solutions. Similar works on this path can also be found in~\cite{baldassi2015subdominant,baldassi2016unreasonable}, where
the discrete networks are considered.

However, the existing research only considered the small difference between the good solutions, which are not addressing the two key issues we
described previously. In~\cite{zhang2016understanding}, through several numerical experiments, the authors suggest that both explicit and implicit
regularizers, when well tuned, could help to reduce the generalization error to a small margin. However, it is far away from the essence of interpreting generalization ability.

\section{Deep neural networks learn low-complexity solutions}
In general, supervised learning usually involves an optimization process of minimizing the empirical risk, $\RR_{emp}(f) := \frac{1}{N} \sum_{i=1}^N \ell(f(\bx_i), y_i)$,
\begin{equation}
 \hat{f} = \argmin_{f \in \HH } \RR_{emp}(f),
\end{equation}
where $\{ (\bx_i, y_i)\}_{i=1}^N$ denotes the training set with $N$ \emph{i.i.d.} samples, $\ell(\cdot, \cdot)$ is the loss function; $\HH$ denotes the whole hypothesis space, and the hypothesis $f$ is often parameterized as $f(\bx; \bth)$, such as deep neural networks. According to central limit theorem (CLT), the generalization error (i.e. the population version of risk) of \emph{one particular learned model} $\hat{f}$, $\RR(\hat{f}) =\bE_{\bx,y} [\ell(\hat{f}(\bx),y)]$, can be decomposed into two terms
\begin{equation}
	\RR(\hat{f}) \leq \RR_{emp}(\hat{f}) +  \sqrt{\var_{\bx,y}\ell(\hat{f}(\bx),y)}/\sqrt{N},
	\label{eqn: hypothsis-generalization-error}
\end{equation}
where the last term is closely related to the complexity of the solution $\hat{f}$, i.e. the complexity of input-output mapping $\hat{f}$.
So with the same and small training error, simple solutions generalize better than complex solutions. This intuitive explanation is called
Occam's razor, and No Free Lunch theorem~\cite{shalev2014understanding} and also related to the minimum description length (MDL) theory~\cite{rissanen2007information,myung2000importance}.

\subsection{Optimizers converge to low-complexity solutions}
\begin{figure}[!ht]
\begin{center}
\includegraphics[width=0.37\textwidth]{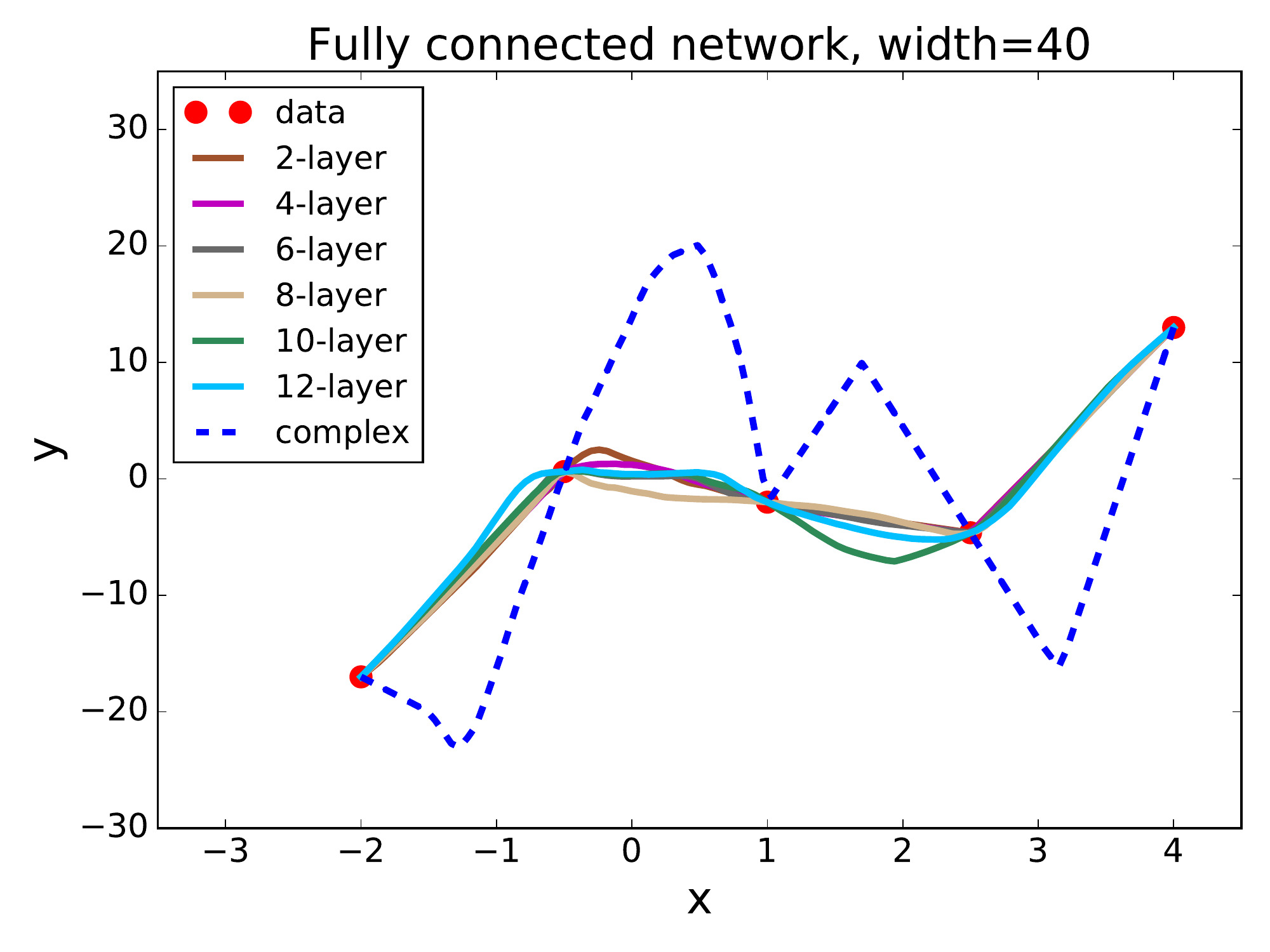}
\hspace{0.5cm}
\includegraphics[width=0.37\textwidth]{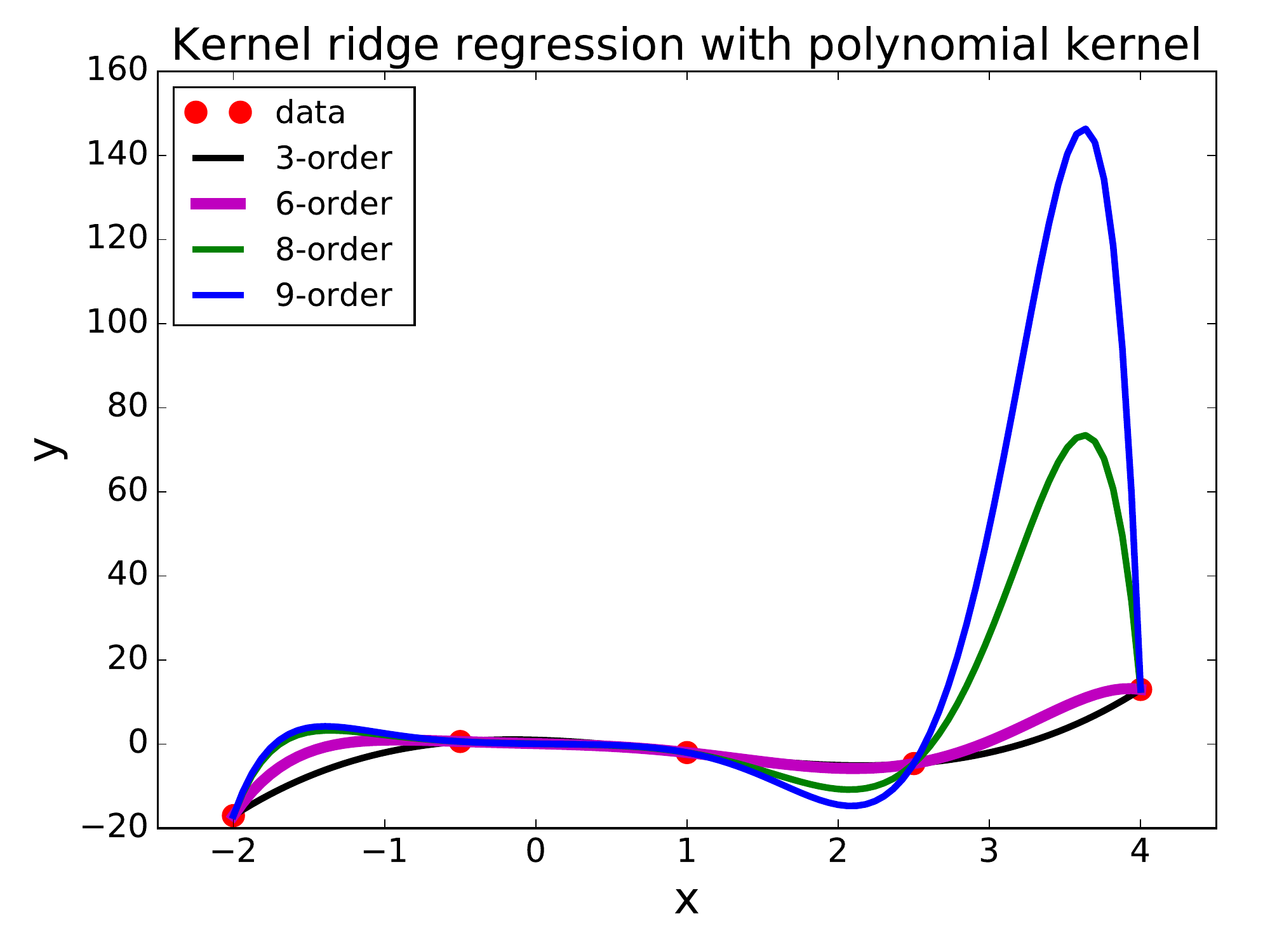}
\end{center}
\vspace{-0.4cm}
\caption{\small{ \text{\bf (Left), } fitting results for 5 data points using FNNs with different number of layers; the overfitting solution with a high complexity (in dashed line) is intentionally constructed.
\text{\bf (Right), } fitting results by kernel regression with different orders of polynomial kernels.}}
\label{fig: 1dexample}
\end{figure}

In deep learning, $\RR_{emp}$ can always be trained to an ignorable threshold $\epsilon$. So it is the complexity of candidate solutions that determine the generalization error according to the bound \eqref{eqn: hypothsis-generalization-error}. To get some intuition about this, we use
fully connected neural networks (FNN)\footnote{ReLU is used as our activation
function through all the experiments in this paper} with different number of  layers to fit a  three-order polynomial: $ y = x^3 -3 x^2 - x +1 +\mathcal{N}(0,0.1)$.
As a comparison, we also conduct the experiment using kernel ridge regression (KRR) model. In this experiment, training set consists of only $5$ points.
For neural networks, all the regularizers are removed; and the regularization parameter for KRR is set to $0.01$. The result
is summarized in Figure~\ref{fig: 1dexample}.
we can easily observe that the optimizer did converge  to the solutions with
low complexity for FNNs with different number of layers. Especially, the $12$-layer network still generalizes well, even with about $18000$ parameters that is much larger than the size of training set. Maybe one thinks that it is because the hypothesis
space of FNN is not as complicated as we imagine. However, this is not true;
and we can find many high-complexity solutions, one of them shown as the dash line in Figure~\ref{fig: 1dexample}. The overfitting solution in the figure can be found by attacking the training set intentionally, see Section~\ref{sect: contrsuct-minima} for more details. On the other hand, KRR inevitably produces overfitting solutions when increasing the capacity. To control the complexity of the solutions for KRR models, we have to resort to some regularization.

\subsection{Connection with classical learning theory}
Classic statistical learning theory describes the generalization error bound for the hypothesis space $\HH$ as follows (we present it in a non-rigorous way for simplicity),
\begin{equation}
	\RR(f) \leq \RR_{emp}(f) + \text{complexity}(\HH)/\sqrt{N}, \quad \forall f \in \HH,
	\label{eqn: ged}
\end{equation}
where the term complexity$(\HH)$ measure the complexity of the whole hypothesis class $\HH$,
 such as VC dimension~\cite{vapnik1998statistical} and Rademacher complexity~\cite{bartlett2002rademacher,zhang2016understanding}, etc. In spite of similarity with the bound~\eqref{eqn: hypothsis-generalization-error}, we emphasize that this bound is universal for the whole hypothesis space $\HH$, and even holds for the worst solution. Therefore, it is not surprising that $\text{complexity}(\HH)$
only leads to  a trivial upper bound~\cite{zhang2016understanding}. However, in practice, what we really care about is the complexity of the specific solutions
found by the optimizers, not the worst case.

As shown in Figure~\ref{fig: 1dexample} and Section~\ref{sect: contrsuct-minima}, for high-capacity deep neural networks,
 the solution set $\HH^*:=\{ f \,|\, \RR_{emp}(f) < \epsilon, \text{ with } \epsilon \text{ small enough.}\}$ consists of many solutions  with diverse generalization performance,  some of which even generalize no better than a random guess. Surprisingly,  optimizers with random initialization rarely converge to the bad solutions in $\HH^*$.
As a comparison, traditional basis expansion methods do not have the nice property of converging to low-complexity solutions (see the  analysis in Supplementary Materials).
Thus, conventional learning theory cannot answer the \textbf{Question 2}.

\section{Perspective of loss landscape for understanding generalization }
The key point to understand the generalization capability of high-capacity deep learning models boils down to figuring out the  mechanism that guides the optimizers to converge towards low-complexity solution areas,  rather than
pursue tighter bounds for complexity of hypothesis class $\HH$ and the solution set $\HH^*$.
Recalling the optimization dynamics of SGD, there exists only three factors that might endow the deep learning models with good generalization performance,
	(1) the stochasticity introduced by mini-batch approximation;
	(2) the specific initialization;
	(3) the special structure of the landscape of $\RR_{emp}(\bth)$.
 After systematical investigation over the three factors, our central finding can be summarized as follows,
 \begin{center}
 \emph{ The geometrical structure of loss function of deep neural networks guides the optimizers to converge to low-complexity solutions: the volume of basin of good minina dominates over those poor ones such that random initialization induces  starting parameters located in good basins with an overwhelming probability, leading to the almost sure convergence to good solutions. }
 \end{center}

\subsection{SGD is not the magic}
\label{sec:sgd}
The stochastic approximation of the gradient was originally proposed to
overcome the computational bottleneck in classic gradient descent. It seems natural to imagine that the noise
introduced by the stochastic approximation should help the system to
escape the saddle points and local minima. Also some researchers suggest that
SGD acts as an implicit regularizer guiding the system to converge to a small-norm/better solutions~\cite{zhang2016understanding,hardt2015train}.

To evaluate these observations, we trained three networks by batch gradient
descent and SGD on MNIST dataset. Here we only use the first  $10,000$ images as the
training set, due to the limited RAM of GPU for the evaluation of full batch gradient. The architectures of the networks
are described in the Supplementary Materials. As shown in Table~\ref{tab: fullbatch}, the
models trained by SGD do perform better than full batch gradient descent,
which is similar as the observed results in \cite{keskar2016large}. However this
improvement is limited. We thus can conclude that SGD alone cannot explain the good generalization capability of deep neural networks, since the generalization we consider focuses on the significant difference between the good solutions with above $95\%$ test accuracy and those poor ones not better than a random guess.
\begin{table}[!ht]
\centering
\caption{\small{The test accuracy (``mean$\pm$std'') of the obtained solutions under different networks, based on 6 runs of the experiments.} }
\begin{tabular}{c|ccc}
\hline
	model         &  mLeNet            &   LeNet            &   sNIN~\cite{lin2013NIN} \\
\hline
	full batch ($10^5$)   &  $97.75\pm 0.24 $ &   $97.85 \pm 0.05$ & $96.89\pm 0.41$  \\
	mini batch ($128$) &  $98.52\pm 0.13$ &   $97.87\pm 0.28$  & $98.41 \pm 0.17$\\
\hline
\end{tabular}
\vspace{-0.3cm}
\label{tab: fullbatch}
\end{table}
\vspace{-0.6cm}

\subsection{The role of landscape and random initialization}
\label{sec:lands_and_init}
To describe the characteristics of the loss landscape, we borrow the concept of
\textit{basin of attraction}~\cite{smaleODE} from dynamical systems, which is the region such that any initial point in
that region will eventually be iterated into the attractor (i.e. minima), denoted as $\CA$.
Let $\BB_{\good},\BB_{\bad}$ be the basins of attractor $\CA_{\good}$
and $\CA_{\bad}$ w.r.t. the optimization dynamics $\dot{\bth} = - \nabla_{\bth} \RR_{emp}(\bth)$, respectively. The empirical observation (see Table~\ref{tab: random}) that random initialization $\bth_0 \sim \mathbb{P}_0$ converges
to $\CA_{\good}$, indicates that
\begin{equation}
    \mathbb{P}_0(\bth_0 \in \mathcal{B}_{\good}) \approx 1,\quad
    		 \mathbb{P}_0(\bth_0 \in \BB_{\bad}) \approx 0 .
    \label{eqn: summary}
\end{equation}
If we choose $\mathbb{P}_0$ to be the uniform distribution, then we have
\begin{equation}
    \frac{\vol(\BB_{\bad})}{\vol(\BB_{\good})} \approx 0,
    \label{eqn: ratio}
\end{equation}
where $\vol(\BB)$ denotes the volume of the basin of attractor.
In terms of Lebesgue measure, the basin of bad minima is a zero measure set.
According to \eqref{eqn: ratio}, a random initialization of parameters will be  located in the basin of good minima $\CA_{\good}$ with an overwhelming probability.
Consequently, optimizers will converge to the solutions generalizing well almost surely.
So we conjecture that the ratio \eqref{eqn: ratio} should be the reasonable answer to \textbf{Question 2}  in  Section~\ref{sec:intro}.
Now we empirically demonstrate that some  random initialization  will result in convergence to good solutions.


We numerically tried different strategies of random initialization, $\bth_0 \sim \mathcal{U}[0,0.1]$, $\mathcal{U}[0,1]$, $\NN(0,0.1)$,  and the one proposed in~\cite{he2015delving}, i.e. $  \NN(0,2/\text{fan}_{in})$, where $\text{fan}_{in}$ the number of inputs for each node.   No data augmentation or regularization is applied in this experiment. For each network and method of random initialization, we run the experiments 6 times.
The results are reported in Table~\ref{tab: random}. It can be easily observed that, with these strategies of random initialization, all the cases converge to good solutions. This partially supports our conjecture that random initialization induces that the starting parameters are almost surely located in the basin of good minima.

\begin{table}[!ht]
\centering
\begin{tabular}{c|c |c }
\hline Initialization 		&  	LeNet (MNIST)	&  ResNet-18 (CIFAR10) \\
\hline
 $\NN\left(0,2/\text{fan}_{in}\right)$&   99.92 $\pm$ 0.15 /  99.00 $\pm$ 0.19 & 100.00 $\pm$ 0.00 /  84.48 $\pm$ 0.20\\
\hline
$\mathcal{U}[0,1]$ 					  & 99.82 $\pm$ 0.33 /  98.99 $\pm$ 0.23 & 100.00 $\pm$ 0.00 /  79.06 $\pm$ 0.59 \\
$\mathcal{U}[0,0.1]$	 			  & 99.97 $\pm$ 0.01 /  99.19 $\pm$ 0.03&100.00 $\pm$ 0.00 /  81.54 $\pm$ 0.28	\\
$\NN(0,0.1)$            			  & 99.97 $\pm$ 0.01 /  99.13 $\pm$ 0.11 &100.00 $\pm$ 0.00 /  84.56 $\pm$ 0.40\\
\hline
\end{tabular}
\vspace{-0.3cm}
\caption{\small{Training/test accuracies for the models trained with various initialization strategies.
	For each experiment, we preprocess the data by channel-wise normalization and train it with SGD optimizer for
	fixed number of epochs, 50 for LeNet (MNIST) and 160 for ResNet-18 (CIFAR10).
}}
\label{tab: random}
\end{table}

\vspace{-0.5cm}
\section{Landscape of 2-layer networks}
\label{sec:2layer}
In this section, we analyze the landscape of 2-layer networks to show that the low-complexity solutions are indeed located in ``flat'' regions with large volume of attractor basin.

The hypothesis  represented by the 2-layer network can be written as,
\begin{equation}
    f(\bx;\bth) := \sum_{k=1}^K a_k \sigma(\vb_k^T \bx + c_k ),
    \label{eqn: 1-hidden-network}
\end{equation}
where $K$ is the number of hidden nodes, $\sigma(\cdot)$ denotes the activation function, and $\bth=(\va,\vc,\vb_1,\cdots,\vb_k)$ denotes all the parameters. Assume least square loss is used, then fitting  $\{(\bx_i,\by_i)\}_{i=1}^N$ becomes minimizing
$
     \RR_{emp}(\bth) := \frac{1}{N}\sum_{i=1}^N \left(f(\bx_i;\bth)-y_i\right)^2.
    \label{eqn: relu2}
$
Here the Hessian matrix of $\RR_{emp}(\bth)$ can be decomposed into two terms:
Fisher information and fitting residual,
\begin{equation}
        \nabla_{\bth}^2 \RR_{emp}(\bth) = \frac{1}{N}\sum_{i=1}^N \nabla_{\bth} f(\bx_i;\bth) \nabla_{\bth}f(\bx_i;\bth)^T + \frac{1}{N}\sum_{i=1}^N \left(f(\bx_i)-y_i\right) \nabla_{\bth}^2 f(\bx_i;\bth).
        \label{eqn: hessian}
\end{equation}
The first term in the R.H.S. of Eq.~\eqref{eqn: hessian} is the empirical Fisher information matrix.
The corresponding population version is defined as $I_{\bth}=\mathbb{E}_{\bx}[\nabla_{\bth}f \nabla_{\bth}f^T]$.
According to ~\eqref{eqn: 1-hidden-network}, for any $\bx$, we have
\begin{equation}
\begin{aligned}
    \frac{\partial f}{\partial a_k} = \sigma(\vb_k^T \bx + c_k), \quad
    \frac{\partial f}{\partial \vb_k} =
     a_ks_k(\bx)\bx, \quad \frac{\partial f}{\partial c_k} = a_k s_k(\bx), \quad
     \frac{\partial f}{\partial x_l} = \sum_{k=1}^K a_k s_k(\bx) b_{k,l},
\end{aligned}
\label{eqn: relu-gradient}
\end{equation}
where $s_k(\bx) = \sigma'(\vb_k^T\bx + c_k)$. 

To measure the complexity of a hypothesis $f$, we choose
 $\mathbb{E}\|\nabla_{\bx} f\|_2^2$ due to its merit of considering  derivatives w.r.t. input $\bx$, which reflects the spatial fluctuation of $f$ . According to Eq.~\eqref{eqn: relu-gradient}, we have
\begin{equation}
\begin{aligned}
    \mathbb{E}\|\nabla_{\bx} f(\bx)\|_2^2 = \sum_{k_1,k_2}\vb_{k_1}^T\vb_{k_2} I_{\vc}(k_1,k_2),
\end{aligned}
    \label{eqn: complexity}
\end{equation}
where $I_{\vc}$ is the Fisher information matrix w.r.t model parameters $\vc$ .
By Cauchy-Schwarz inequality, we can obtain the following theorem to relate the complexity of hypothesis with the Fisher information matrix w.r.t. model parameters, see Supplementary Materials for the proof.
\begin{theorem}
\label{them: 1}
For any $f(x;\bth)$ expressed by a 2-layer network according to Eq.~\eqref{eqn: 1-hidden-network}, let $B=(\vb_1,\dots,\vb_K)\in \R^{d\times K}$, then  we
have
\begin{equation}
	2 \mathbb{E}\|\nabla_{\bx}f\|_2^2 \leq \|B\|_F^4 +  \|I_{\vc}\|_F^2.
\end{equation}
\end{theorem}
The above theorem establishes the relationship between the hypothesis complexity measured by the norm of expected input gradient and
the Fisher information matrix. Additionally, the latter is also related to the Hessian of $\RR_{emp}$ according to \eqref{eqn: hessian}, thus we have
the following characterization of the landscape of $\RR_{emp}$.
\begin{corollary}
\label{corollary: 1}
For any $f$ expressed by a 2-layer network, let $C_{\sigma}:=\|\sigma''(z)\|_{\infty}$, then we have
\begin{equation}
	2 \mathbb{E}\|\nabla_{\bx} f(\bx)\|^2 \leq
		  \|\nabla_{\vc}^2\RR_{emp}\|_F^2
		 + \|B\|_F^4  + 2 C_{\sigma}D \sqrt{\RR_{emp}}
      	 + O\left(\sqrt{\var\|\nabla_{\bx} f\|_2^2 / N }\right)
      \label{eqn: complexity-bound}
\end{equation}
where the last term is the Monte Carlo approximation error, $D=\max_{k} |a_k|\|\vb_k\|_2^2$ and $\nabla^2_{\vc} \RR_{emp}$ denotes the Hessian matrix \textit{w.r.t.} $\vc$.
\end{corollary}
The upper bound~\eqref{eqn: complexity-bound} reveals some remarkable properties of the landscape of $\RR_{emp}(\bth)$.
 We can ignore the last term, if the number of training samples is large enough compared to $\var \|\nabla_{\bx} f\|_2^2$.
For ReLU networks, $C_{\sigma}=0$, so the complexity of a small-norm hypothesis is bounded by the Frobenius norm of Hessian $\nabla_{\vc}^2\RR_{emp}$.  For general activation functions, this is true
for the hypotheses with small training error, i.e. $\RR_{emp}$ small enough.

Without loss of generality, we can add constraint $\mathcal{C}:= \{\bth\, | \, \|B\|_F^4 + 2 \max_{k} |a_k|\|\vb_k\|_2^2 \leq \eta\}$, with $\eta$ being very small. Since $f(\bx;\bth)$
is invariant to the \emph{node-scaling}~\cite{neyshabur2015path}, i.e. $(a_k,\vb_k,c_k) \rightarrow (t^{-1} a_k,t \vb_k,t c_k)$, this constraint doesn't shrink
the hypothesis space. It means that any hypothesis $f(\bx;\bth)$ has at least one corresponding $\bth' \in \mathcal{C}$, such that $f(\bx;\bth)=f(\bx;\bth')$.
  For $\bth \in \mathcal{C}$, the bound~\eqref{eqn: complexity-bound} implies that low-complexity solutions lie in the
areas with small Hessian. This indicates that low-complexity solutions are located in the flat and large basins of attractor, while the high-complexity solutions lie in the sharp
and small ones.
Therefore, a  random initialization tends to produce starting parameters located in the basin of good minima with a high probability, giving rise to the almost sure convergence to good minima using gradient-based methods.

In practice, we do not explicitly impose the constraint $\mathcal{C}$. What we do is to randomly initialize the system close to zero, which implicitly results in the optimizer
exploring the landscape in the small vicinity of zero. Within this area, the high-complexity minima  generalizing like random guessing have much smaller attractor basins.
Therefore empirically we never observe that optimizers converge to these bad solutions, even though they do exist in this area.

In~\cite{dinh2017sharp}, the authors argued that the property of Hessian  cannot
be directly applied to explain generalization. The reason to this argument is  that although $\mathbb{E}\|\nabla_{\bx}f\|_2^2$ is invariant to
node-scaling, the Hessian $\nabla_{\bth}^2\RR_{emp}(\bth)$ not. However, in most cases of neural networks,  the learned solutions are close
to zero (i.e. with small norms) due to the near-zero random initialization, and thus the term $\|B\|_F^4  + 2 C_{\sigma}D \sqrt{\RR_{emp}}$ in the bound~\eqref{eqn: complexity-bound} is dominated  by the Hessian  $\|\nabla_{\vc}^2\RR_{emp}\|_F^2$. Therefore, it is reasonable to apply the property of the Hessian to explain the generalization ability.


Our theoretical analysis sheds light on the difference between the minima that generalize well and bad, answering the \textbf{Question 1} raised in Section~\ref{sec:intro}. This part only provides a rough analysis of hypothesis-dependent generalization for 2-layer neural networks.
 For deeper networks, more elaborated analysis is left as future work.


\section{Experiments}
\label{sec:exp}
For deep neural networks, it is  difficult to analytically analyze the landscape of $\RR_{emp}(\bth)$, so we resort to numerical evidence, as shown in this section.
\subsection{Construct solutions with diverse generalization performance}
\label{sect: contrsuct-minima}
To numerically demonstrate the property described in Eq.~\eqref{eqn: ratio} and the bound~\eqref{eqn: complexity-bound}, we need find a large number of minima with diverse generalization ability, particularly including the solutions that perform as nearly random guesses in test set. However, this is
not easy if only relying on the training set, since the random initialization always converges to solutions generalizing well, due to Eq.~\eqref{eqn: ratio}. To overcome this difficulty, we
design an extra attack dataset to ``fool'' the networks to produce bad generalization performance.

For each model, we prepare three datasets, $\CS_{\text{\bf train}},\CS_{\text{\bf test}},\CS_{\text{\bf attack}}$, representing the training, test  and attack set, respectively.
All the data points on the attack set are intentionally assigned wrong labels. Then we solve the following optimization problem instead of the original one,
\begin{equation}
	\min_{\bth } \RR^{a}_{emp}(\bth) := \underbrace{\frac{1}{|\CS_{\train}|}\sum_{\bx\in \CS_{\text{\bf train}}} \ell(f(\bx;\bth),y)}_{\RR_{\text{\bf train}}(\bth) = \RR_{emp}(\bth)} +
                        \gamma  \underbrace{\frac{1}{|\CS_{\attack}|}\sum_{\bx\in \CS_{\text{\bf attack}}}\ell(f(\bx;\bth),y_{\text{wrong}})}_{\RR_{\text{\bf attack}}(\bth)}.
	\label{eqn: attack}
\end{equation}
Because of the high capacity of over-parameterized neural networks, we can obtain various solutions achieving $ \RR^{a}_{emp}(\bth^*)\leq \varepsilon$, thus $\RR_{\text{\bf train}}(\bth^*)\leq \varepsilon$.
But due to the attack term, the performance on the test set is harmed severely. In practice  by tuning the hyperparameter $\gamma$ and the size of attack set,
 a series  of solutions of $\RR_{\text{\bf train}}(\bth)$ can be found, and their generalization error can be very bad.

\subsection{Spectral analysis of Hessian matrices}
Since the volume of attractor basin is a global quantity, it is hard to estimate directly.
Fortunately, a large basin often implies that the local valley around the
 attractor is very flat and vice versa. Similar ideas were also explored in~\cite{hochreiter1997flat,chaudhari2016entropy,keskar2016large}. However, their numerical
 experiments only consider solutions from $\CA_{\good}$. We are investigating the difference between $\CA_{\good}$ and $\CA_{\bad}$ to understand why optimizers with random
 initialization rarely converges to $\CA_{\bad}$.
\begin{figure}[!h]
\begin{center}
\includegraphics[width=0.32\textwidth]{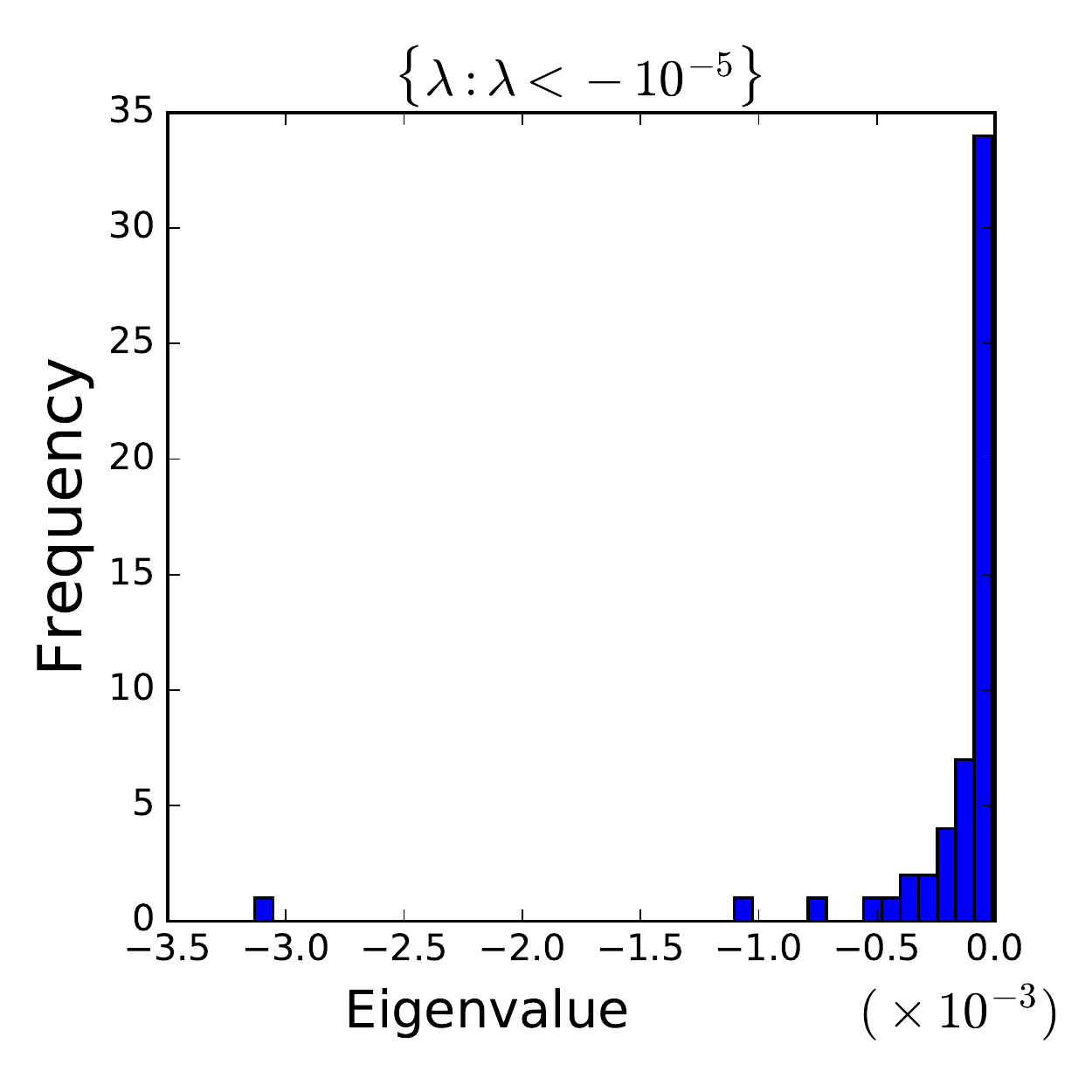}
\includegraphics[width=0.32\textwidth]{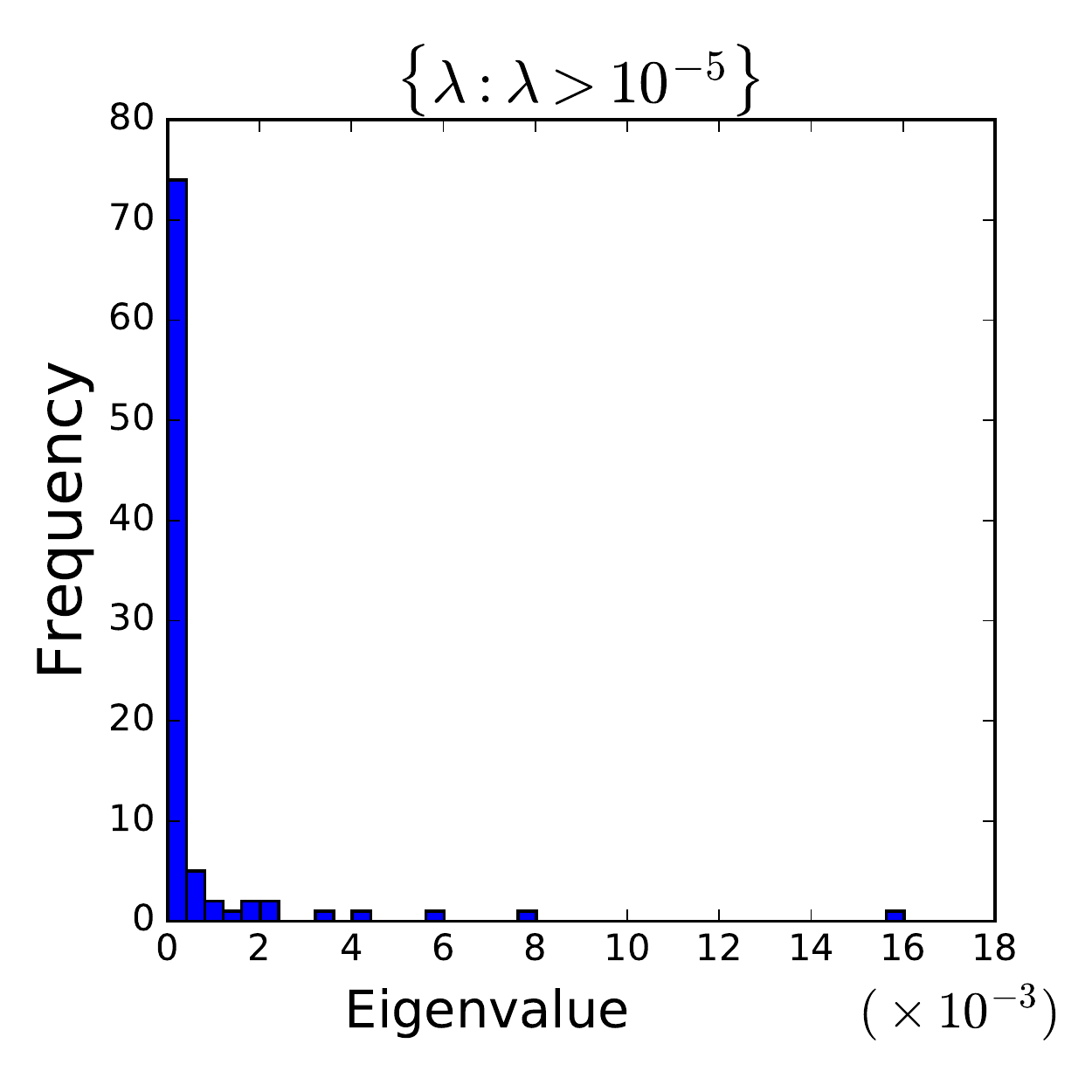}
\includegraphics[width=0.32\textwidth]{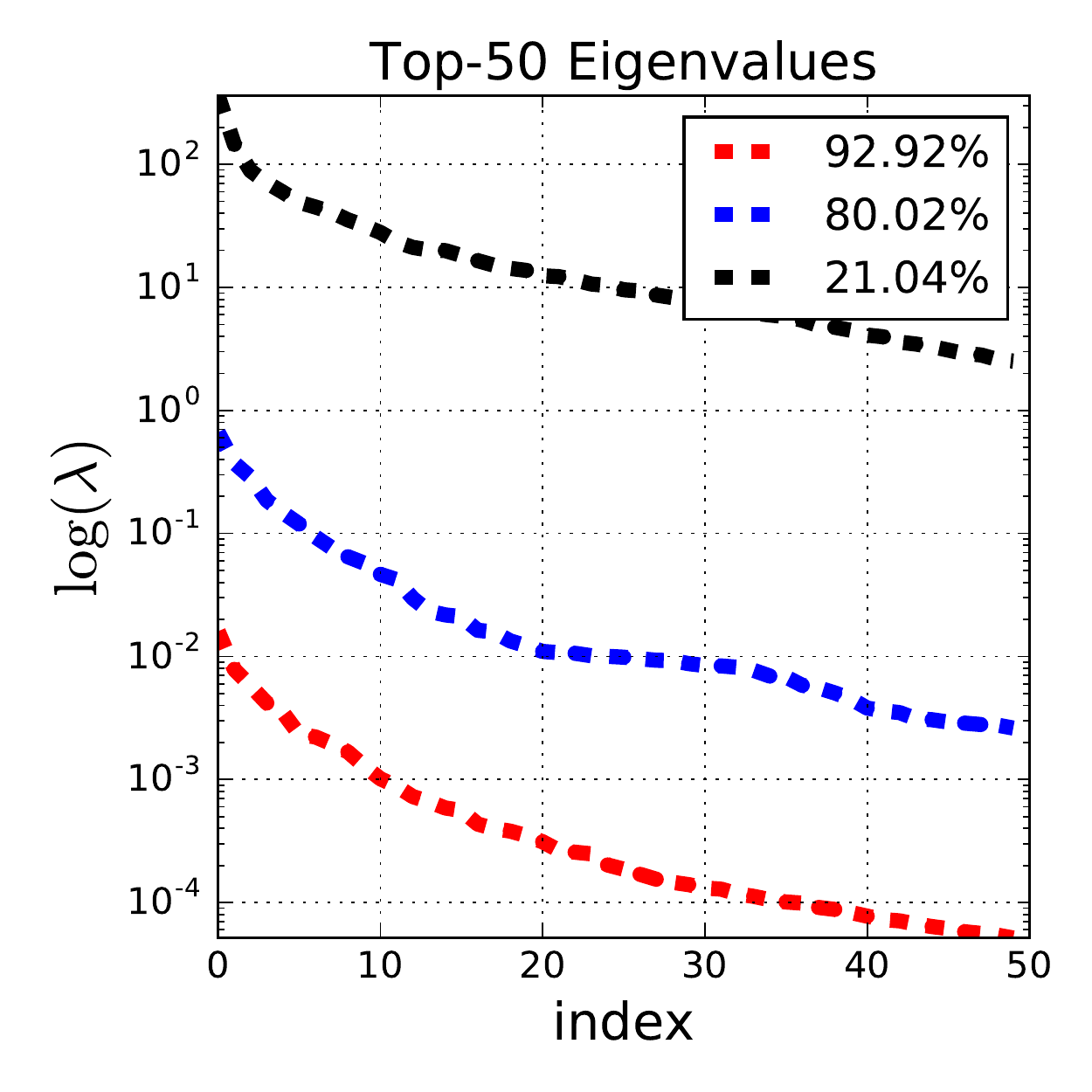}
\end{center}
\caption{\small{\textbf{(Left, Middle)} Eigenvalues distribution of a model with $92\%$ test accuracy; \textbf{(Right)} Top-$k$ eigenvalues for three solutions, all with
    training accuracy $100\%$. The model used here is mLeNet (number of parameter is 3781), and dataset is MNIST. In the experiment, the first $512$ training data are selected as our
    new training set with the rest of training data as  attack set. The model is initialized by $\mathcal{N}(0,2/\text{fan}_{in})$. 
    } }
\label{fig: spectral-mnist}
\vspace{-0.5cm}
\end{figure}
\vspace{-0.2cm}

Figure~\ref{fig: spectral-mnist} shows an example of the spectrum of the Hessian of a small CNN around minima. The following observations  are not unique to this model, shared across different
models and datasets~\cite{chaudhari2016entropy,sagun2016singularity}. (1) There are some negative eigenvalues since the optimizer is terminated before it
converges to the strict minima. (2) \emph{Most of the eigenvalues concentrate around zero}. In accordance to the work~\cite{freeman2016topology}, { the good solutions
form a connected manifold (allowing small energy barrier). We conjecture that, the large amount of zero eigenvalues might imply that the dimension of this manifold is  large, and the
eigenvectors of the zero eigenvalues span the tangent space of the attractor manifold. The eigenvectors of the other large eigenvalues correspond to the directions away
from the attractor manifold.}
We leave the justification of this conjecture as future work. (3) The right most plot shows that the bad solutions have much larger eigenvalues than the good ones. This indicates that the good solutions
lie in the wide valley while bad solutions sit in the narrow one. It is consistent with our analysis on 2-layer networks.

Based on the above analysis of the spectrum of Hessian, it is natural to use the product of  top-$k$ positive eigenvalues to quantify the inverse volume of the attractor. For a given Hessian matrix $H$ around a solution, we utilize the logarithm of the product of  top-$k$ eigenvalues to approximate  the inverse volume of basin of attractor:
\begin{equation}
       V(k) := \sum_{i=1}^k \log(\lambda_i(H)).
        \label{eqn: volume}
\end{equation}

\subsection{Numerical evidence for deep neural networks}
\paragraph{Small neural networks}
We train dozens of mLeNets on two datasets, MNIST and SVHN.  For each experiment, the first $512$   training data are used as our new training set, while the other
training data are used as attack set to help generate diverse solutions according to~\eqref{eqn: attack}. Different optimizers are adopted to increase the diversity of the solutions.
\vspace{-0.3cm}
\paragraph{Large neural networks}
The ResNet-32 is used to fit CIFAR-10 dataset, where
the first $10^4$ samples of training set are selected as our new training data,
with the remaining as the attack set. The performance is evaluated
on the whole test set. No regularization or data augmentation is used. This model has about $0.5$ million of parameters, which is much
larger than the number of training set, $10^4$. Due to the prohibitive cost to compute the spectrum of Hessian, we employ a statistical estimate
of the Frobenius norm of Hessian, although not perfect but computationally
feasible for large networks.

For any matrix $A\in \R^{q\times p}$, the Frobenius norm can be estimated by
$
	\|A\|_F^2 = \mathbb{E}_{\bv} \|A\bv\|_2^2, \quad \bv \sim \mathcal{N}(0,I).
$
Therefore in practice, after replacing the matrix-vector product with finite
difference, we have the following estimator:
\begin{equation}
	\|H(\bth)\|_F^2 \approx \frac{1}{M}\sum_{i=1}^M \|H(\bth)\bv^i\|^2=\frac{1}{4 M \varepsilon^2}\sum_{i=1}^M\|  \nabla f(\bth + \varepsilon \bv^i) -
						\nabla f(\bth - \varepsilon \bv^i)\|_2^2, \label{eq:hessiannorm}
\end{equation}
where $\{ \bv^i \}_{i=1}^M$ are i.i.d random samples drawn from $N(0,I)$. This estimation of the squared Frobenius norm of Hessian becomes exact when $\varepsilon\rightarrow 0$ and $M\rightarrow + \infty$. In this experiment, we set $M=100, \varepsilon=10^{-5}$.

\begin{figure}[!th]
\vspace{-0.1cm}
\begin{center}
\includegraphics[width=0.32\textwidth]{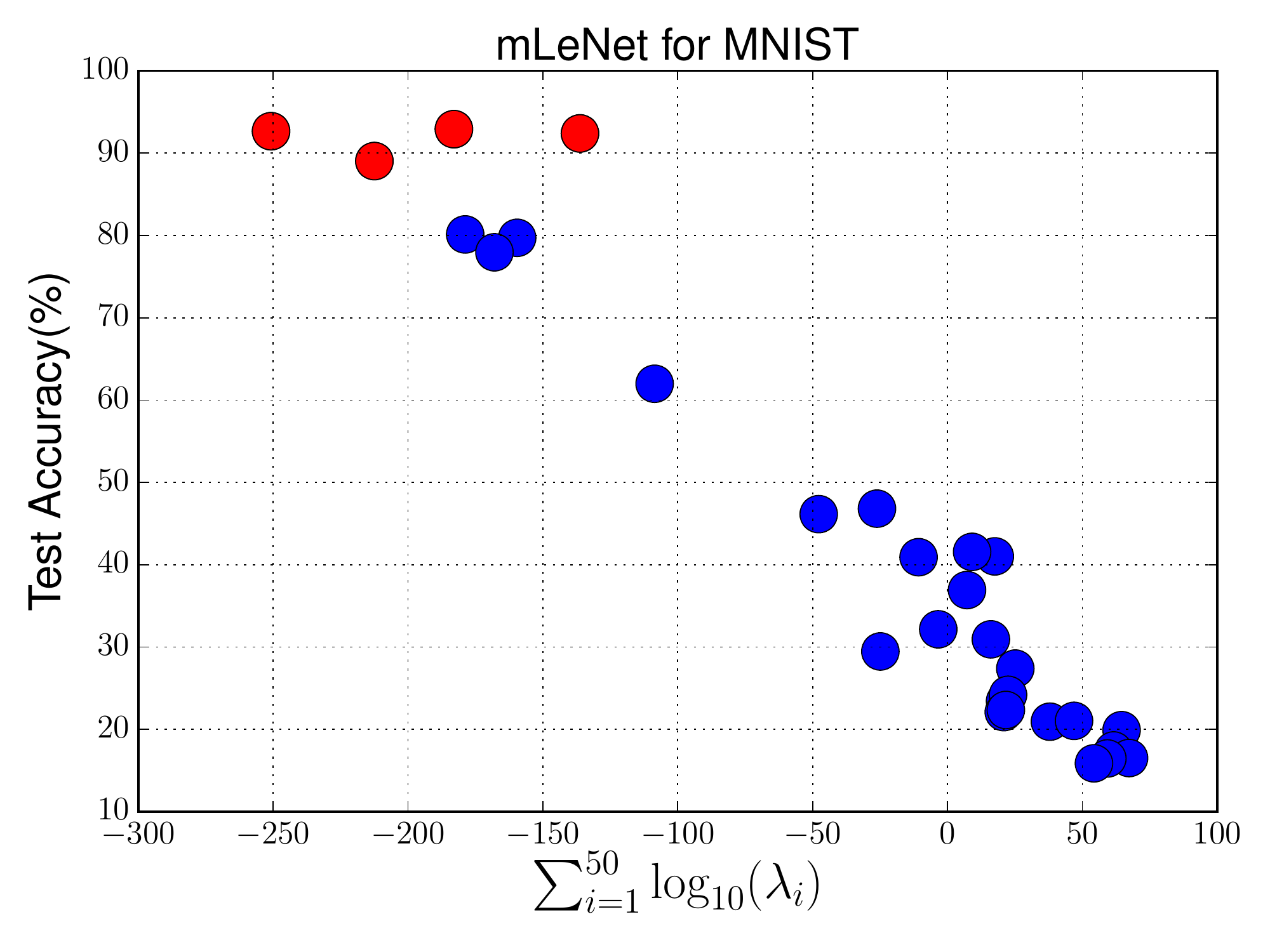}
\includegraphics[width=0.32\textwidth]{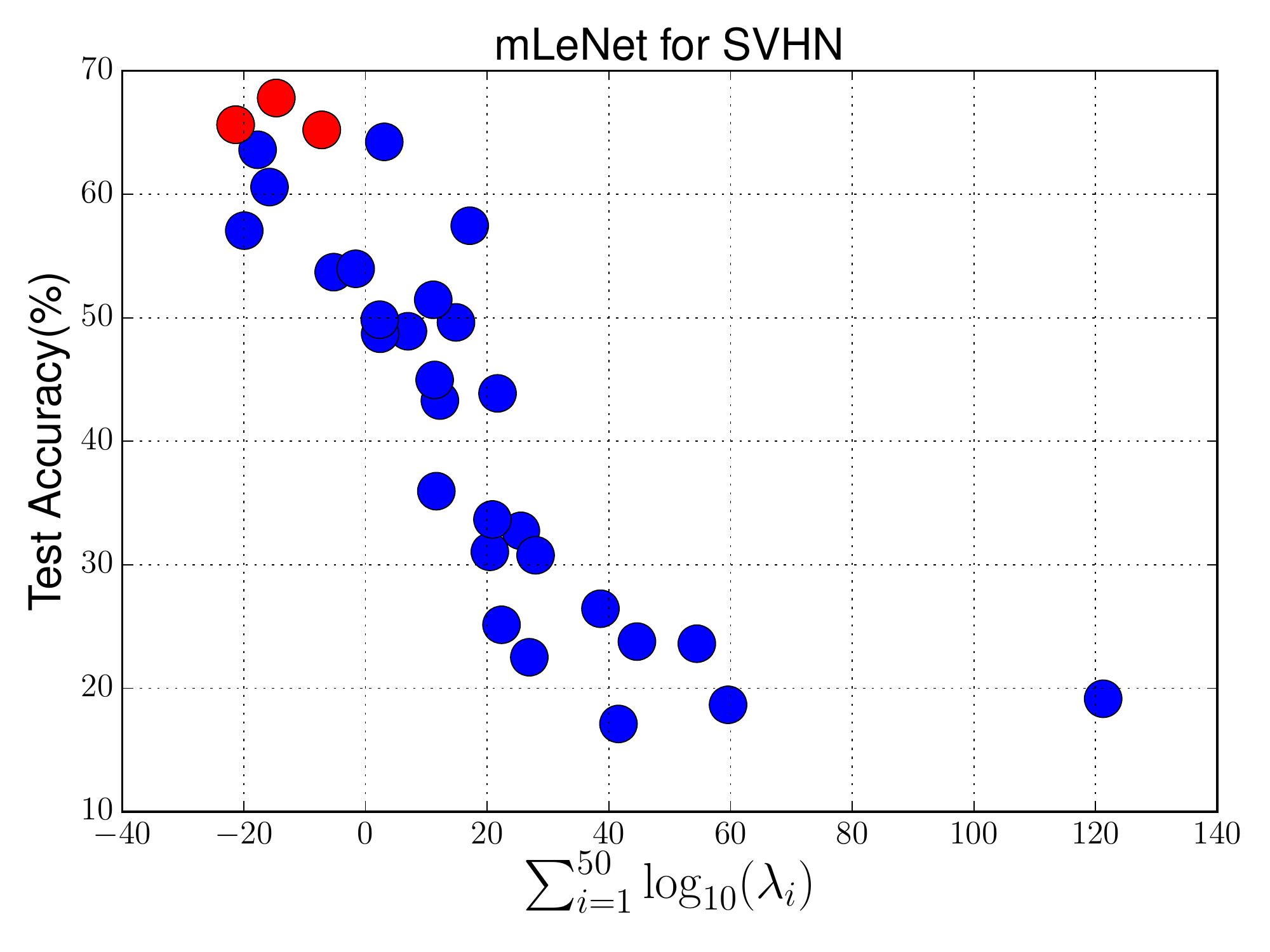}
\includegraphics[width=0.32\textwidth]{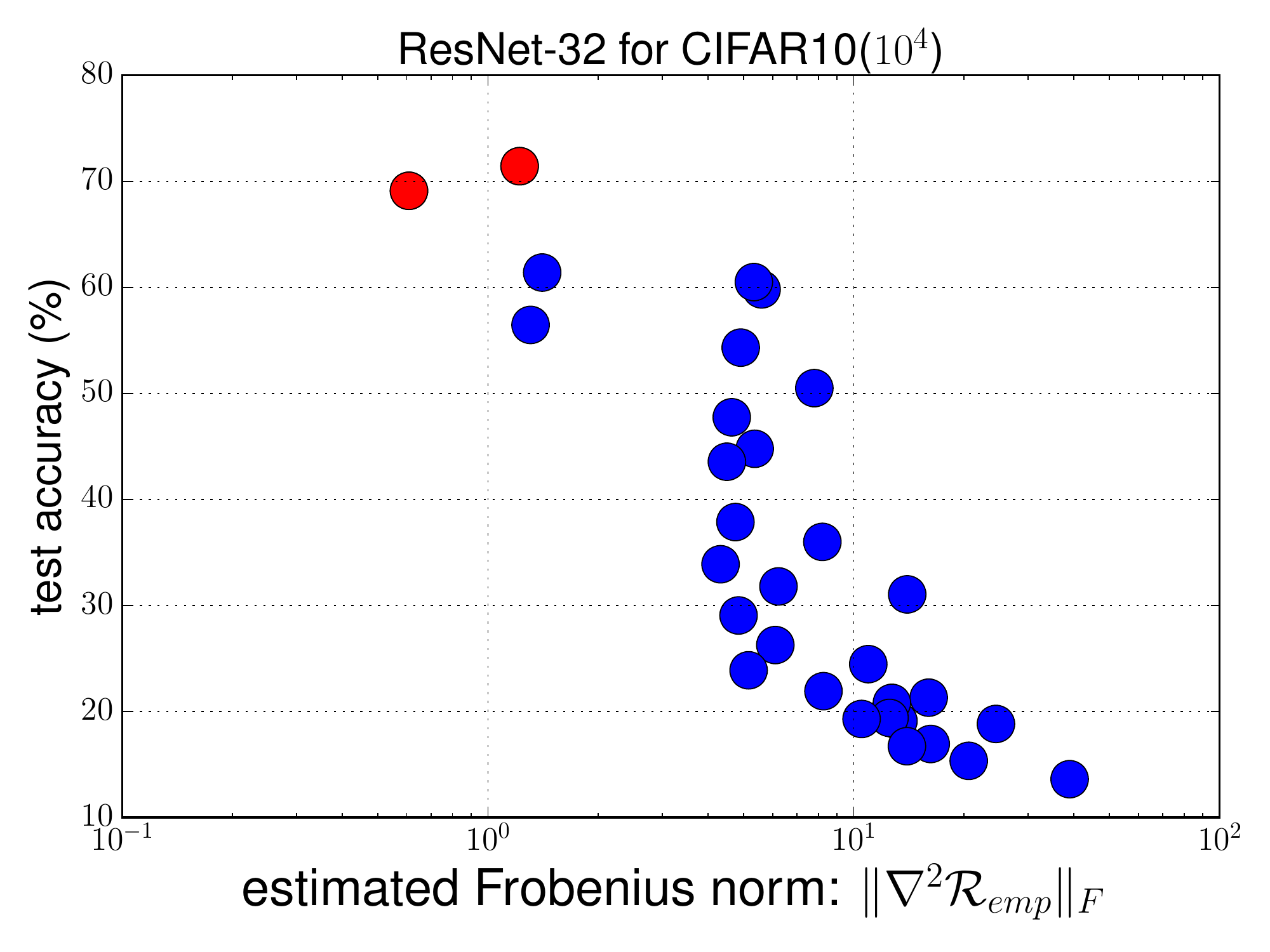}
\end{center}
\vspace{-0.5cm}
\caption{\small{Test accuracies versus volume of basin (approximated by different quantities).
The red bubbles represent the solutions found by optimizers without the attacking term. All solutions are initialized from $\mathcal{N}(0,2/\text{fan}_{in})$ and trained to achieve about $100\%$ accuracy on training set.}
}
\label{fig: scatter-cnn}
\end{figure}
The numerical results for both small and large networks are shown
in Figure~\ref{fig: scatter-cnn} to reveal the relationship between the test accuracy and the inverse volume of basins.
We can easily observe that,
\begin{itemize}
\item Good minima are located in very large valley, while bad ones sit in small valley. The ratio $\vol(\BB_{\bad})/\vol(\BB_{\good})$ in Eq.~\eqref{eqn: ratio} through the estimation by Eq.~\eqref{eqn: volume} and~\eqref{eq:hessiannorm}
        can be exponentially small due to the high dimensionality. This evidently supports our findings that the volume of basin of good minima dominates over those generalizing poorly, leading to that the optimization methods with random initialization converge to good solutions almost surely.
\item There exists some variance of the relationship between generalization error and volume of basins. Also it is almost impossible to use the $V(k)$ in Eq.~\eqref{eqn: volume} to distinguish the minima with equivalent generalization performance.  We conjecture that the reason might be that Hessian-based characterization of the volume of basin is only a rough estimate.  A better non-local quantity is necessary.
\end{itemize}

\section{Conclusion}
In this work, we attempt to answer two important questions towards understanding generalization of deep learning: what is the difference between the minima that generalize well and poorly; and why training methods converge to good minima with an overwhelming probability.
The 2-layer networks are analyzed to show that the low-complexity solutions have a small norm of Hessian matrix w.r.t. model parameters. This directly reveals the difference between good and bad minima. We also investigate this property for deeper neural networks through various numerical experiments, though theoretical justification is still a challenge.
This property of the Hessian implies that the volume of basin of good minima dominates over that of poor ones, leading to an almost sure convergence to good solutions, as demonstrated by various empirical results.


\bibliographystyle{plain}
\bibliography{deep-neural-networks}

\begin{thebibliography}{10}

\bibitem{baldassi2016unreasonable}
C.~Baldassi, C.~Borgs, J.~Chayes, A.~Ingrosso, C.~Lucibello, L.~Saglietti, and
  R.~Zecchina.
\newblock Unreasonable effectiveness of learning neural networks: From
  accessible states and robust ensembles to basic algorithmic schemes.
\newblock {\em Proceedings of the National Academy of Sciences},
  113(48):E7655--E7662, 2016.

\bibitem{baldassi2015subdominant}
C.~Baldassi, A.~Ingrosso, C.~Lucibello, L.~Saglietti, and R.~Zecchina.
\newblock Subdominant dense clusters allow for simple learning and high
  computational performance in neural networks with discrete synapses.
\newblock {\em Physical review letters}, 115(12):128101, 2015.

\bibitem{bartlett2002rademacher}
P.~L Bartlett and S.~Mendelson.
\newblock Rademacher and gaussian complexities: Risk bounds and structural
  results.
\newblock {\em Journal of Machine Learning Research}, 3(Nov):463--482, 2002.

\bibitem{chaudhari2016entropy}
P.~Chaudhari, A.~Choromanska, S.~Soatto, and Y.~LeCun.
\newblock Entropy-sgd: Biasing gradient descent into wide valleys.
\newblock {\em In International Conference on Learning Representations (ICLR)},
  2017.

\bibitem{dinh2017sharp}
L.~Dinh, R.~Pascanu, S.~Bengio, and Y.~Bengio.
\newblock Sharp minima can generalize for deep nets.
\newblock {\em arXiv preprint arXiv:1703.04933}, 2017.

\bibitem{freeman2016topology}
J.~Freeman, C. D.and~Bruna.
\newblock Topology and geometry of half-rectified network optimization.
\newblock {\em In International Conference on Learning Representations (ICLR)},
  2017.

\bibitem{hardt2015train}
M.~Hardt, B.~Recht, and Y.~Singer.
\newblock Train faster, generalize better: Stability of stochastic gradient
  descent.
\newblock In {\em Proceedings of The 33rd International Conference on Machine
  Learning}, pages 1225--1234, 2016.

\bibitem{he2015delving}
K.~M. He, X.~Y. Zhang, S.~Q. Ren, and J.~Sun.
\newblock Delving deep into rectifiers: Surpassing human-level performance on
  imagenet classification.
\newblock In {\em Proceedings of the IEEE international conference on computer
  vision}, pages 1026--1034, 2015.

\bibitem{smaleODE}
M.~W. Hirsch, S.~Smale, and R.~L. Devaney.
\newblock {\em Differential equations, dynamical systems, and an introduction
  to chaos}.
\newblock Academic press, 2012.

\bibitem{hochreiter1997flat}
S.~Hochreiter and J.~Schmidhuber.
\newblock Flat minima.
\newblock {\em Neural Computation}, 9(1):1--42, 1997.

\bibitem{hochreiter1995simplifying}
S.~Hochreiter, J.~Schmidhuber, et~al.
\newblock Simplifying neural nets by discovering flat minima.
\newblock {\em Advances in Neural Information Processing Systems}, pages
  529--536, 1995.

\bibitem{keskar2016large}
N.~S. Keskar, D.~Mudigere, J.~Nocedal, M.~Smelyanskiy, and P.~T.~P. Tang.
\newblock On large-batch training for deep learning: Generalization gap and
  sharp minima.
\newblock In {\em In International Conference on Learning Representations
  (ICLR)}, 2017.

\bibitem{lecun2015deep}
Y.~LeCun, Y.~Bengio, and G.~Hinton.
\newblock Deep learning.
\newblock {\em Nature}, 521(7553):436--444, 2015.

\bibitem{lin2013NIN}
M.~Lin, Q.~Chen, and S.C. Yan.
\newblock Network in network.
\newblock {\em arXiv preprint arXiv:1312.4400}, 2013.

\bibitem{myung2000importance}
I.~J. Myung.
\newblock The importance of complexity in model selection.
\newblock {\em Journal of mathematical psychology}, 44(1):190--204, 2000.

\bibitem{vapnik1998statistical}
Vladimir N.~Vapnik.
\newblock {\em Statistical learning theory}, volume~1.
\newblock Wiley New York, 1998.

\bibitem{neyshabur2015path}
B.~Neyshabur, R.~R. Salakhutdinov, and N.~Srebro.
\newblock Path-sgd: Path-normalized optimization in deep neural networks.
\newblock In {\em Advances in Neural Information Processing Systems}, pages
  2422--2430, 2015.

\bibitem{rissanen2007information}
J.~Rissanen.
\newblock {\em Information and complexity in statistical modeling}.
\newblock Springer Science \& Business Media, 2007.

\bibitem{sagun2016singularity}
L.~Sagun, L.~Bottou, and Y.~LeCun.
\newblock Singularity of the hessian in deep learning.
\newblock {\em arXiv preprint arXiv:1611.07476}, 2016.

\bibitem{shalev2014understanding}
S.~Shalev-Shwartz and S.~Ben-David.
\newblock {\em Understanding machine learning: From theory to algorithms}.
\newblock Cambridge university press, 2014.

\bibitem{ye2017langevin}
N.~Y. Ye, Z.~X. Zhu, and R.~K Mantiuk.
\newblock Langevin dynamics with continuous tempering for high-dimensional
  non-convex optimization.
\newblock {\em arXiv preprint arXiv:1703.04379}, 2017.

\bibitem{zhang2016understanding}
C.~Zhang, S.~Bengio, M.~Hardt, B.~Recht, and O.~Vinyals.
\newblock Understanding deep learning requires rethinking generalization.
\newblock In {\em International Conference on Learning Representations}, 2017.

\end{thebibliography}


\section*{Supplementary Materials}
\subsection*{Appendix A. Model details}
This following list gives the details of the model used in this paper:
\begin{description}
	\item[LeNet]  Standard model.
    \item[mLeNet] A modified LeNet. This model is usually used to compute the Hessian.
     				Since we need to compute a large number of Hessian matrices,
     				the last fully connected layer of LeNet is  replaced by   a convolutional 
     				layer plus a global average pooling layer. The number of model parameter
 					is $\bf{3781}$.
  \item[sNIN] A small network in network model.  This model is used to conduct the full batch gradient descent experiments. Because of the limited GPU memory, the feature numbers of three block of NIN is set to $(20,30,20)$.
  \item[ResNet-18/ResNet-32] Standard model (batch normalization is used).
\end{description}

\subsection*{Appendix B: landscape of convex model}
For many shallow models, such as basis function modeling and kernel methods,
the good and bad solutions are not distinguishable if only relying on the information of empirical risk $\RR_{emp}(\bth)$. For instance, considering the following model,
\begin{equation*}
	\RR_{emp}(\bth) = \frac{1}{N} \sum_{i=1}^N \ell\left(\bth^T \bm{\phi}(\bx_i), y_i \right),
	\label{eqn: bfm2}
\end{equation*}
where $\bth \in \R^d$, $d>N$, $\bm{\phi}(\bx)$ is the feature map. The second-order derivatives can be written as
\begin{equation*}
	\nabla_{\bth}^2 \RR_{emp} = \frac{1}{N}\sum_{i=1}^N \partial_{\hat{y}_i}^2\ell(\hat{y}_i,y_i) \bm{\phi}(\bx_i)  \bm{\phi}(\bx_i)^T, \quad \hat{y}_i =\bth^T \bm{\phi}(\bx_i)
\end{equation*}
For most loss functions used in practice, $\partial_{\hat{y}}^2\ell(\hat{y},y)$ is a
constant at $\hat{y}=y$, e.g. $L_2$ loss, hinge loss, etc. It implies that
around the global minima $\bth^*$, the loss surface has both the same first-order and second-order information, i.e. $\nabla_{\bth}\RR_{emp} = 0$ and $\nabla_{\bth}^2 \RR_{emp} = \text{constant}$.  This directly leads to that optimizers themselves are unable to find the solution generalizing better  due to the indistinguishability between different global minima. Thus, in order to steer the optimizers to find the low-complexity solutions,  the only way is to introduce some proper regularization to shrink the hypothesis space.

 Different from other models, the loss surface of deep neural networks owns the  minima that are distinguishable with each other, see detailed analysis in Section~\ref{sec:lands_and_init} and \ref{sec:2layer}. The possibility stems from the non-convexity of neural networks, which make it possible to seperate the
 good and bad solutions into different valleys, distinguished via the information of $\RR_{emp}$.

\subsection*{Appendeix C: Theoretical result proof details}
\subsubsection*{Proof of theorem \ref{them: 1}}
Because $\frac{\partial f}{\partial x_l} = \sum_{k=1}^K a_k s_k(\bx) b_{k,l}$ and $  \frac{\partial f}{\partial c_{k}} = a_k s_k(\bx)$, so
\begin{equation}
\begin{aligned}
    \mathbb{E}\|\nabla_{\bx} f(\bx)\|^2 &=
        \int \rho(d\bx) \sum_{l} \sum_{k_1,k_2} a_{k_1} a_{k_1}\vb_{k_1,l}\vb_{k_2,l} s_{k_1}(\bx)s_{k_2}(\bx) \\
                        & =  \int \rho(d\bx) \sum_{k_1,k_2} a_{k_1} a_{k_1}\vb_{k_1}^T\vb_{k_2} s_{k_1}(\bx)s_{k_2}(\bx)  \\
                        & = \sum_{k_1,k_2}\vb_{k_1}^T\vb_{k_2} \mathbb{E} \left [ \frac{\partial f}{\partial c_{k_1}}  \frac{\partial f}{\partial c_{k_2}}  \right ] \\
                        & = \sum_{k_1,k_2}\vb_{k_1}^T\vb_{k_2} I_{\vc}(k_1,k_2) \\
\end{aligned}
\end{equation}
Thus
\begin{equation}
    \begin{aligned}
         \mathbb{E}\|\nabla_{\bx} f(\bx)\|^2 &\leq \half \sum_{k_1,k_2} \left(\vb_{k_1}^T \vb_{k_2}\right)^2 + \frac{1}{2} I^2_{\vc}(k_1,k_2) \\
                    &\leq \half \sum_{k_1,k_2} \|\vb_{k_1}\|^2 \|\vb_{k_2}\|^2 + \half \sum_{k_1,k_2}I^2_{\vc}(k_1,k_2) \\
                    &= \half \left( \sum_{k} \|\vb_k\|^2_2\right)^2 +
                     \half \|I_{\vc}\|_F^2
    \end{aligned}
\end{equation}

\subsubsection*{Proof of corollary \ref{corollary: 1}}
According to central limit theorem, we have
\begin{equation}
\begin{aligned}
    \mathbb{E}\|\nabla_{\bx} f(\bx)\|^2  & = \sum_{k_1,k_2}\vb_{k_1}^T\vb_{k_2} I_{\vc}(k_1,k_2) \\
        &\approx  \sum_{k_1,k_2}\vb_{k_1}^T\vb_{k_2} I^N_{\vc}(k_1,k_2)  + O\left(\sqrt{\frac{\var\|\nabla_{\bx} f\|_2^2 }{N }}\right)\\
      &= \sum_{k_1,k_2}\vb_{k_1}^T\vb_{k_2} \left(\frac{\partial^2 \RR_{emp}}{\partial {c_{k_1}}\partial {c_{k_2}}}-  \frac{1}{N}\sum_{i=1}^N (f(\bx_i)-y_i)\frac{\partial^2 f}{\partial {c_{k_1}}\partial {c_{k_2}}} \right) + O\left(\sqrt{\frac{\var\|\nabla_{\bx} f\|_2^2 }{N }}\right).
\end{aligned}
\end{equation}
Since $ \frac{\partial^2 f}{\partial {c_{k_1}}\partial {c_{k_2}}} = \delta_{k_1,k_2}
 \sigma''(\vb_{k_1}^T \bx + c_{k_1})$, we have
\begin{equation}
\begin{aligned}
      \mathbb{E} \|\nabla_{\bx} f(\bx)\|^2 & \leq \half \left( \sum_{k} \|\vb_k\|^2_2\right)^2 + \half \|\nabla_{\vc}\RR_{emp}\|_F^2 \\
      & \qquad -\underbrace{\frac{1}{N}\sum_{k_1,k_2,i} (f(\bx_i)-y_i) \vb_{k_1}^T\vb_{k_2} \delta_{k_1,k_2}a_{k_1} \sigma''(\vb_{k_1}^T \bx_i + c)}_{Q} + O\left(\sqrt{\frac{\var\|\nabla_{\bx} f\|_2^2 }{N }}\right).
\end{aligned}
\label{eqn: app1}
\end{equation}
Next we assume the second-order derivative of activation function is  bounded, i.e. $C_{\sigma}:=\|\sigma''(x)\|_{\infty} $ is finite. This assumption is satisfied by commonly-used activation functions, like sigmoid, tanh and ReLU, etc. Especially for ReLU, $C_{\sigma}=0$. So we have,
\begin{equation}
\begin{aligned}
Q &= \frac{1}{N}\sum_{k}  \left(f(\bx_i)-y_i\right) \|\vb_k\|^2 a_k \sigma''(\vb_k^T\bx + c_k)\\
 & \leq \frac{1}{N}\sum_{k} |f(\bx_i)-y_i| \|\vb_k\|^2 a_k C_{\sigma}\\
 & \leq C_{\sigma} D \sqrt{\RR_{emp}}
\end{aligned}
\label{eqn: app2}
\end{equation}
where $D=\max_{k} |\|\vb_k\|^2a_k|$. Combining Eq.~\ref{eqn: app1} and Eq.~\ref{eqn: app2}, we obtain the following characterization of $\RR_{emp}$,
\begin{equation}
	2 \mathbb{E}\|\nabla_{\bx} f(\bx)\|^2 \leq \|\nabla_{\vc}^2\RR_{emp}\|_F^2
				+ \|B\|_F^4  + 2 C_{\sigma}D  \sqrt{\RR_{emp}}
      	 + O\left(\sqrt{\var\|\nabla_{\bx} f\|_2^2 / N }\right).
\end{equation}

\end{document}